# Implementing Binary Transformers with Approximate Memories with Permutation Invariance Enabled Selective (PIES) Protection

Zhen Gao, *Senior Member, IEEE*, Xiao Yin, Kexin Liang, Pedro Reviriego, *Senior Member*, *IEEE*, Shanshan Liu, *Senior Member*, *IEEE*, Fabrizio Lombardi, *Life Fellow*, *IEEE*

***Abstract*—Quantization is an effective technique to reduce memory storage, computation complexity and power for Transformers, and the binarization of the weights is important for meeting these requirements. Approximate memory can further reduce the power for memory, but it also introduces errors. Current research for binary Transformers focuses on task performance, but the model dependable operation against errors on the memory has not been well studied. In this paper, we first evaluate the dependability of binary Transformers with the Bidirectional Encoder Representations from Transformers (BERT) as a case study. A large accuracy loss occurs only at high Bit Error Rates (e.g. BER = 5% or 10%). Then we propose a novel Permutation Invariance Enabled Selective (PIES) scheme to protect the binary weights at very low overhead. This scheme exploits the invariance of Transformers to permutations in the rows and columns to group the critical weights and perform selective protection on those weights. Simulation results show that PIES can recover most of the performance loss caused by bit errors at BERs as high as 10%, at an overhead significantly lower than traditional error correction schemes. These results confirm the feasibility of running binary Transformers using approximate memory with aggressive voltage scaling on resource-limited low power edge devices. The code for PIES is released on github (https://github.com/YangZhutian/binary-transformer-prefix-repair).**

***Index Terms*—Binary Transformers, BERT, Dependability, Fault Tolerance, Permutation Invariance**

## I. Introduction

The introduction of the Transformer architecture by Google in 2017 has marked a milestone in artificial intelligence, particularly for Natural Language Processing (NLP) and computer vision [1]-[6]. As a widely applied Transformer model, Bidirectional Encoder Representations from Transformers (BERT) has shown very good performance in text classification, question answering systems, and sentiment analysis. However, BERT models typically need a massive number of parameters and demand substantial computational resources and power, so they pose significant challenges for deployment on resource-constrained devices. To address this issue, various model compression techniques, including pruning, knowledge distillation, and quantization, have been proposed in technical literature. Among them, binarization as an extreme form of quantization can compress model weights to merely 1 bit, drastically reducing storage requirements and inference latency. Moreover, approximate memories can be implemented for both SRAMs and DRAMs by reducing the supply voltage or refresh frequency [7], [8]. This technique can significantly reduce the energy dissipation at the cost of introducing errors in the memory bits, thus making the memory approximate [9]. The impact of these errors depends on the application and the importance of the bits that suffer the errors [10]; approximate memories have been proposed for video applications [11], data sketches [12] or deep learning implementations [13] showing the potential of energy dissipation reduction for data storage and thus, for the overall computing system.

For approximate memories, one of the key issues is to evaluate the dependability of the models to soft errors. The dependability of Transformers has been initially studied in the last few years. In [14], the dependability of BERT to soft errors in floating-point weights has been evaluated. The results show that errors on a critical bit significantly affect model performance. Furthermore, [15] and [16] have examined the dependability of BERT models and Large Language Models (LLMs) against errors on quantized weights based on error injection. These simulation results have shown that quantized weights are more dependable to soft errors than floating-point weights, and both models operate reliably at high Bit Error Rates (BERs). In particular, the high significant bits are much more vulnerable to errors than lower bits.

For the protection of weights in Transformers against memory errors, several selective protection schemes have been proposed. For example, [14] has found that the first exponent bit of weights in some Transformers is always 0, so we can mitigate the impact of bit flip errors by directly setting the bit back to 0. Sparrow has proposed to apply error correction code (ECC) only for the several Most Significant Bits (MSBs) [17]. Similarly, SHIFT has applied stronger ECC for higher bits and weaker ECC for lower bits [18]. Furthermore, [19] has proposed to protect the quantized KV cache by copying the two MSBs to the two Least Significant Bits (LSBs); this also applies to quantized weights. In addition, the authors in [20] have proposed to improve the dependability of binary neural networks by maximizing the inference margin; however, this approach relies on specially designed training objectives.

As far as the authors know, although existing studies indicate that binarized BERT models achieve reasonable

Zhen Gao, Xiao Yin and Kexin Liang are with the School of Electrical and Information Engineering, Tianjin University, Tianjin 300072, China. Email: {zgao, yin_xiao, lkxx}@tju.edu.cn.
Pedro Reviriego is with the Information Processing and Telecommunications Center, ETSI de Telecomunicación, Universidad Politécnica de Madrid, Madrid 28040, Email: pedro.reviriego@upm.es.
Shanshan Liu is with the School of Information and Communication Engineering, University of Electronic Science and Technology of China, Chengdu, Sichuan 611731, China. Email: ssliu@uestc.edu.cn.
Fabrizio Lombardi is with Department of Electrical and Computer Engineering, Northeastern University, Boston, MA 02215, USA. Email: lombardi@ece.neu.edu.

accuracy with very low computational complexity [21], [22], their dependability to soft errors has not been studied. In particular, the selective protection for MSBs of weights does not apply for binary weights. In this paper, we evaluate the dependability of two typical binarized BERT models against soft errors and propose a novel method to efficiently protect part of the critical binary weights. The main contributions of the paper are summarized as follows:

- We perform a complete dependability evaluation of binary BERTs to bit flip errors; this has revealed that the models are very dependable to errors, and we only need to protect the binary weights for BER higher than 5%;
- We apply a Genetic Algorithm (GA) to find the critical weights for dependability of binary BERTs, and show that only protecting a small portion of them (e.g. 5%) can effectively improve the dependability of the models;
- We propose a Permutation Invariance Enabled Selective protection scheme (PIES) to gather the critical weights in the first few rows or columns of the weight matrices, and show that protecting the gathered weights introduces a negligible performance loss;
- We apply two methods to implement the PIES scheme (selective TMR and Block ECC) and show the feasibility of PIES to achieve a higher dependable performance at a lower cost than traditional ECC schemes.

The remainder of this paper is organized as follows: Section II provides preliminaries for the paper, including an overview of the two binarized BERT models, related tasks and the permutation invariance in Transformers. Section III introduces the dependability evaluation setup and reports the evaluation results. Section IV explores the possibility of selective protection based on GA search. Section V shows the details of the PIES scheme and Section VI presents the implementation of PIES based on TMR and BECC and the evaluation results. Finally, Section VII ends the paper with key considerations.

## II. Preliminaries

In this section, we first introduce two well-known binary BERTs (BinaryBERT [21] and BIT [22]) in subsection A and B, respectively, and then introduce the tasks used in our evaluation and the performance of the two models in subsection C. The basics of permutation invariance in Transformers is finally discussed in subsection D.

### *A. Introduction of BinaryBERT*

The BinaryBERT is generated by de-composing a pre-trained Ternary BERT (as shown in Fig. 1). In this section, we first give a brief introduction of the Ternary BERT and then describe how to split the ternary weights to binary weights. Finally, the quantization of the activations is introduced.

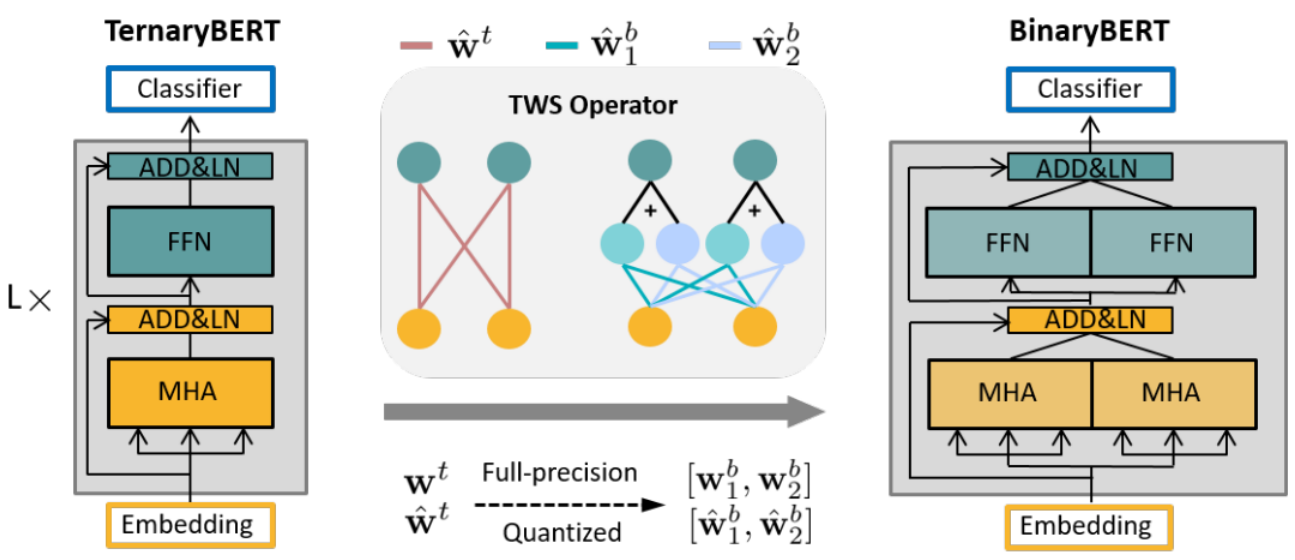

Fig. 1. From Ternary BERT to Binary BERT [21]

#### *1) Ternary BERT*

The Dynamic BERT (DynaBERT) [23] is used as the backbone for the training of Ternary BERT. The DynaBERT we select in this paper has the same architecture as BERT-base, it consists of 12 Transformer encoders, and each encoder includes 12 heads. The hidden dimension is 768, and the intermediate size for FFN is 3072. The sizes of the different types of weight matrices are listed in Table I, including $W_{Q/K/V}$ for self-attention (SA) and $W_{F1/F2}$ for the forward feed network (FFN). Each item in a weight matrix is quantized to one of the three values $\{-\alpha, 0, \alpha\}$. For the original weight matrix $\boldsymbol{W^t}$ (with $n$ items), the $i$-th weight is denoted as $w_i^t$, so the ternary quantized weight can be expressed as equation (1), in which $\Delta = \frac{0.7}{n}\|\boldsymbol{W^t}\|_1$ and $\alpha = \frac{1}{|I|}\sum_{i \in I}|w_i^t|$, $I = \{i \,|\, |w_i^t| > \Delta\}$. For the training of the Ternary BERT, knowledge distillation with the guidance of the outputs of the intermediate layers and the final logits are used to improve performance of the model.

$$\hat{w}_i^t = \begin{cases} \alpha \cdot sign(w_i^t) & |w_i^t| \geq \Delta \\ 0 & |w_i^t| < \Delta \end{cases} \tag{1}$$

TABLE I Size of Weight Matrices in Ternary BERT

| Type | $W_Q$ | $W_K$ | $W_V$ | $W_O$ | $W_{F1}$ | $W_{F2}$ |
|---|---|---|---|---|---|---|
| Size | 12×64×768 | 12×64×768 | 12×64×768 | 768×768 | 3072×768 | 768×3072 |

#### *2) Ternary Weight Splitting*

The original weight matrix $\boldsymbol{W^t}$ is divided into two matrices $\boldsymbol{W^t} = \boldsymbol{W_1^b} + \boldsymbol{W_2^b}$, and each item in $\boldsymbol{W_1^b}$ and $\boldsymbol{W_2^b}$ can expressed as equations (2) and (3), respectively. Finally, each weight in $\boldsymbol{W_1^b}$ (or $\boldsymbol{W_2^b}$) is quantized to binary values $\{-\alpha_1, \alpha_1\}$ (or $\{-\alpha_2, \alpha_2\}$) based on equation (4).

$$w_1^b = \begin{cases} a \cdot w_i^t & \text{if } w_i^t \neq 0 \\ b + w_i^t & \text{if } \hat{w}_i^t = 0, w_i^t > 0 \\ b & otherwise \end{cases} \tag{2}$$

$$w_2^b = \begin{cases} (1-a) \cdot w_i^t & \text{if } \hat{w}_i^t \neq 0 \\ -b & \text{if } \hat{w}_i^t = 0, w_i^t > 0 \\ -b + w_i^t & otherwise \end{cases} \tag{3}$$

$$\begin{cases} \hat{w}_1^b = \alpha_1 \cdot sign(w_1^b) \\ \hat{w}_2^b = \alpha_2 \cdot sign(w_2^b) \end{cases}. \tag{4}$$

#### *3) Quantization of Activations*

Since the input token embedding vectors are floating point values, the intermediate results are also floating-point values. In Binary BERT, the intermediate results of each step (activations) are quantized into 8-bit or 4-bit integers. For the results with both positive and negative values, symmetric quantization is applied to each item. This type of results includes the input of the encoder $X$, the Key/Query/Value matrices $K/Q/V = X * W_{K/Q/V} + b_{K/Q/V}$, the outputs of single attention head $SA = V * Attention_score$, and the output of the multi-head attention $MH\text{-}SA = W_O$*Concat($SAs$)+$B_O$. For the results with only positive values, asymmetric quantization is applied. This type of results includes the $Attention_score = softmax(K^TQ)$ and the input of the second fully connection layer of FFN $X_{f2} = activate(O_{f1})$ where $O_{f1}$ is the output of the first fully connection layer and $activate(\cdot)$ is the non-linear activation function.

BinaryBERT applies different activation quantization schemes for 8-bit and 4-bit cases. Uniform quantization is applied for the 8-bit case. For the $i$-th item $x_i$ in an activation vector $X$, the symmetric quantization is performed based on equation (5):

$$q_s(x_i) = s \cdot round\left(\frac{x_i}{s}\right) \quad (5)$$

where $s = max(|X|)/q_{max}$ is the step size, and $q_{max} = 128$ for 8-bit quantization. For asymmetric quantization, the item $x_i$ is first normalized as

$$\tilde{x}_i = \frac{x_i - \beta}{\alpha} \quad (6)$$

in which $\beta = \min(x)$ and $\alpha = \max(x) - \min(x)$, and finally quantized as

$$q_a(x_i) = \frac{\alpha}{q_{max}} round(\tilde{x}_i \cdot q_{max}) + \beta. \quad (7)$$

For the 4-bit case, Learned Step size Quantization (LSQ) is applied. The symmetric quantization has the same form as equation (5), and the asymmetric quantization is performed based on equation (8). In both cases, the step size $s$ becomes a learnable parameter.

$$q_a(x_i) = s \cdot round\left(\frac{x_i - \min(x)}{s}\right) \quad (8)$$

## B. Introduction of BIT

As introduced in [22], BIT also has the same architecture as the BERT-base model. In BIT, all the weights are quantized into binary values, and the activations are also quantized. In this work, the bit-width for the activation quantization can be changed, and the minimum width is 1. Fig. 2 shows the processing procedure for the case of 1-bit quantization for both weights and activations (W1A1).

For a matrix $\boldsymbol{W}$ with $n$ items, the $i$-th item $w_i$ is quantized as per equation (9),

$$\widehat{w}_i = \frac{\|\boldsymbol{W}\|_1}{n} sign(w_i - \overline{w}) \quad (9)$$

in which $\overline{w}$ is the mean value of all weights in $\boldsymbol{W}$. For the activations $x_i$, the 1-bit symmetric and asymmetric quantization schemes are performed based on equations (10) and (11), respectively, where $\alpha$ and $\beta$ are learnable parameters, and $clip(z, 0,1)$ returns 0 when $z < 0.5$ and returns 1 when $z \geq 0.5$.

$$q_s(x_i) = \alpha \cdot sign(x_i - \beta) \quad (10)$$

$$q_a(x_i) = \alpha \cdot clip\left(\frac{x_i - \beta}{\alpha}, 0,1\right) \quad (11)$$

The symmetric and asymmetric activation quantization schemes for larger bit-width (e.g. 4-bit or 8-bit) are the same as for the 4-bit quantization in Binary BERT.

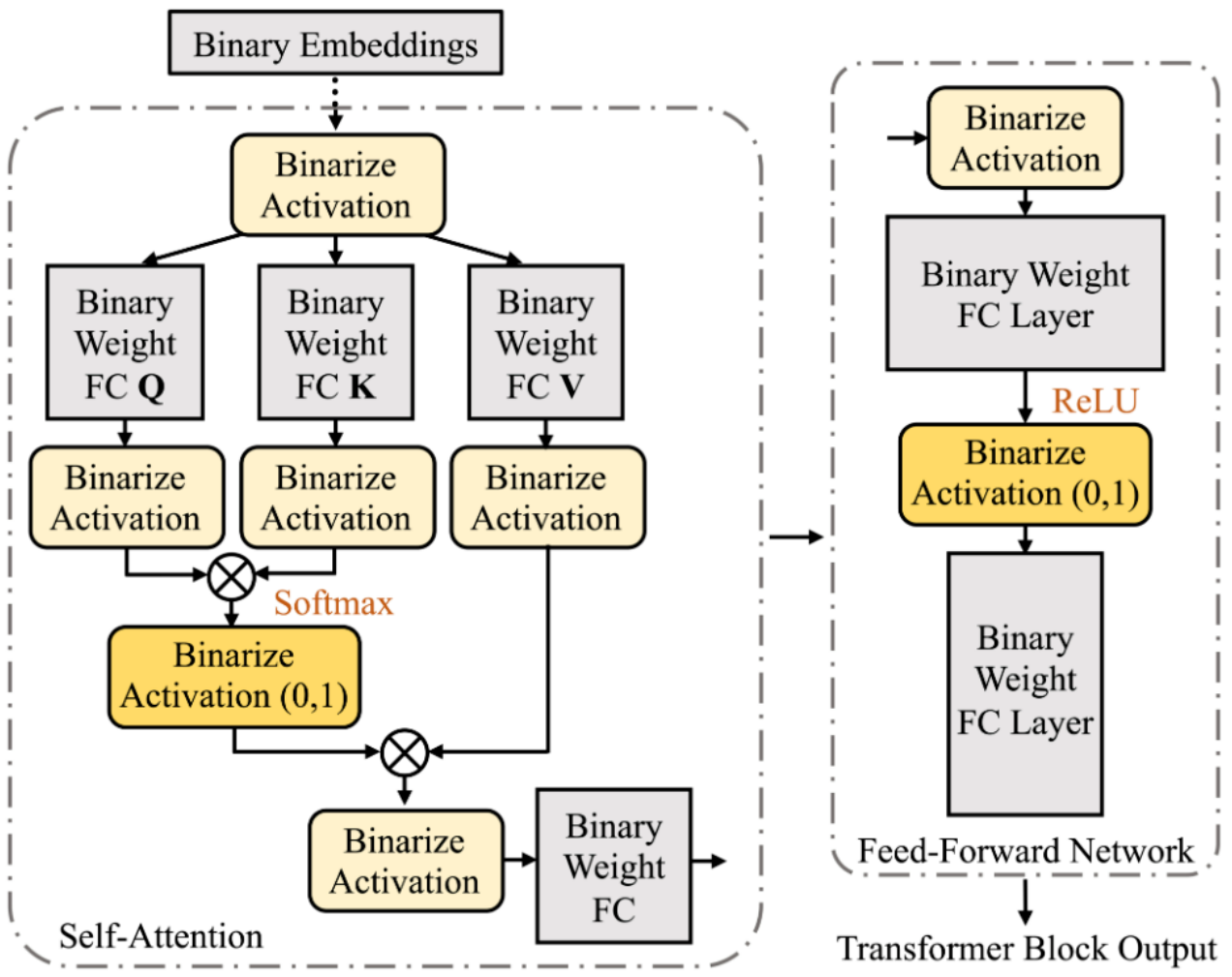

Fig. 2. Processing procedure of BIT for 1-bit weights and activations [22]

## C. Task and Performance for BinaryBERT and BIT

In [21] and [22], the performance of Binary BERT and BIT were both evaluated using the eight tasks from GLUE [24]. In this paper, we select the QNLI (Question Natural Language Inference) dataset from GLUE for the dependability evaluation of Binary BERT (using accuracy as metric). The test set of QNLI includes 5,461 pairs of Question and Answer, and Binary BERT is trained to identify whether the answer is for the question (binary classification task). The open-source code from [25] is used for the evaluation of Binary BERT in this paper. In the error free case, the classification accuracy of BinaryBERT on QNLI is about 91.2% and 90.6% for W1A8 (1-bit weights and 8-bit activations) and W1A4 (1-bit weights and 4-bit activations), respectively, which are very close to the results reported in [21].

For BIT, we selected the SST-2 (The Stanford Sentiment Treebank) from GLUE for the evaluation. The test set of SST-2 includes 1.8k sentences with a binary sentiment label for each sentence (positive or negative). Using the open-source code from [26], the binary classification accuracy of BIT on SST-2 is tested as 85%, 90% and 90% for W1A1, W1A4 and W1A8, respectively, which are also close to the results reported in [22].

## D. Permutation Invariance in Transformers

The processing of Transformers can be divided into three parts: a) Attention score calculation based on Q and K matrix (namely QK circuit); b) Head combination based on V and O matrix (namely OV circuit); c) FFN. As shown in Fig. 3, permutation invariance (PI) holds in all three parts (bias vectors are ignored for simplicity without destroying the conclusion).

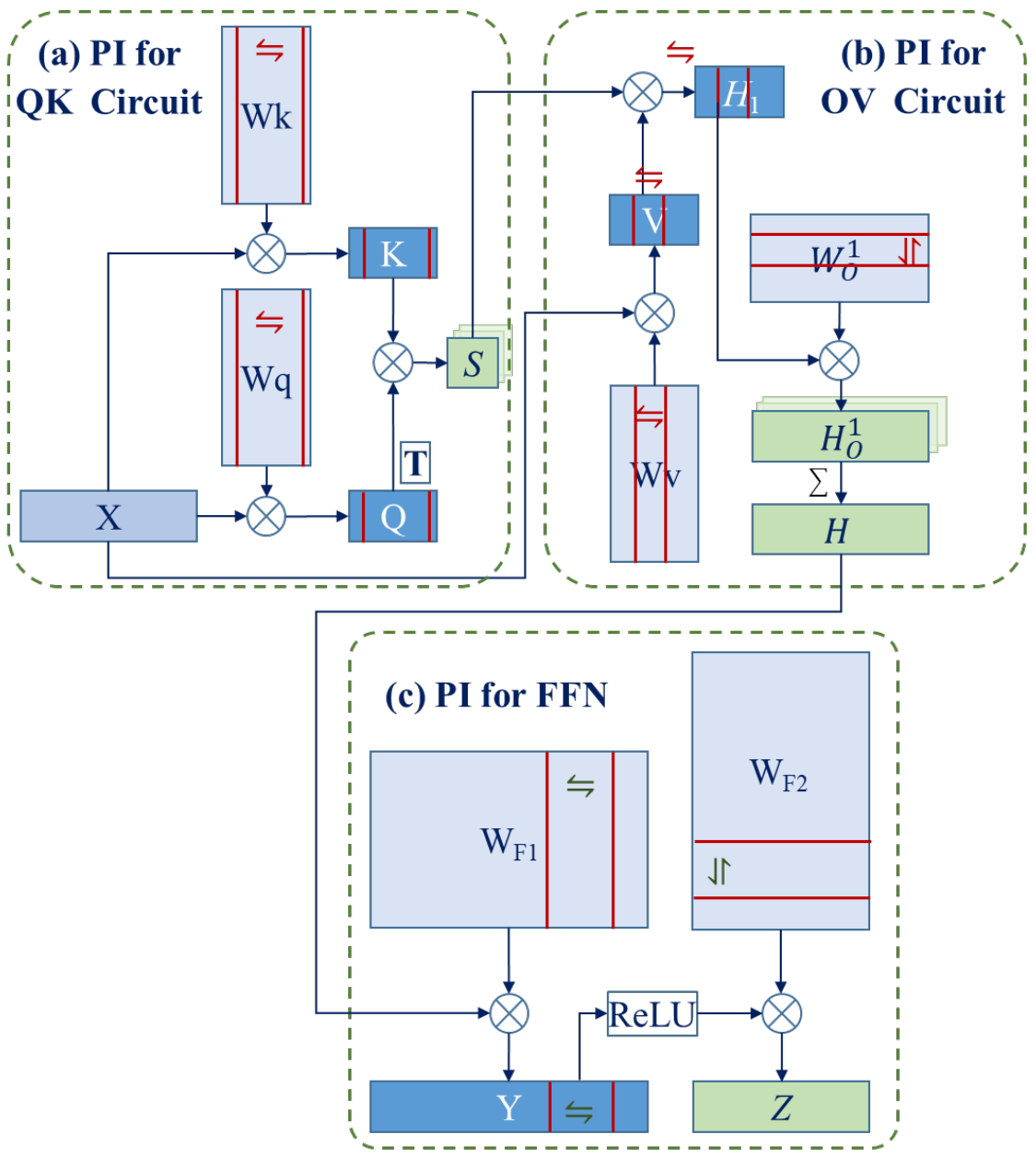

Fig. 3. Permutation invariance in the three parts of Transformers

***PI for QK circuit:*** As shown in Fig. 3(a), for each head, changing the same pair of columns in $W_K$ and $W_Q$ does not change the attention score result S;

***PI for OV circuit:*** The combination of multiple heads can be expressed as $H = \sum_i H_O^i$, in which $H_O^i = H_i \cdot W_O^i$, $H_i = S_i \cdot (W_V^i X)$, $W_O = [W_O^1; W_O^2; \ldots; W_O^N]$. As shown in Fig. 3(b), for the $i$-th head, if we change a pair of columns in $W_V^i$ and the corresponding pair of rows in $W_O^i$, the corresponding result $H_O^i$ and the final $H$ does not change.

***PI for FFN***: As shown in Fig. 3(c), the output of FFN is $Z = W_{F2} \cdot ReLU(W_{F1} \cdot H)$ (by ignoring the bias term); if we change a pair of columns in $W_{F1}$ and the corresponding pair of rows in $W_{F2}$, the result $Z$ does not change.

This permutation invariance property will be exploited for selection protection of critical weights as discussed in Section V.

## III. Dependability Evaluation

In this section, we first introduce the error model considered in sub-section A and then evaluate the impact of the errors on task performance for different error rates and different types of weights in sub-section B and C, respectively.

### *A. Approximate Memories and Error Model*

As mentioned in Section I, using approximate memories for deep learning models can significantly reduce energy consumption but it may increase the occurrence of random bit flip errors in memory. The authors in [27] evaluated the bit error rate (BER) for a 28 nm SRAM with supply voltage scaling. As shown in Fig. 4, 71.6% (46.9%) dynamic (static) power can be saved in the 28 nm SRAM by introducing a BER of $10^{-2}$. Similarly, the authors reported in [28] that the energy per SRAM access can be reduced by 56% when the supply voltage is decreased to 76% of $V_{min}$, and the corresponding BER can be higher than 10% (as shown in Fig. 5). Furthermore, the study for DRAM in [29] also reports that the BER by reducing voltage can also achieve 10% or higher.

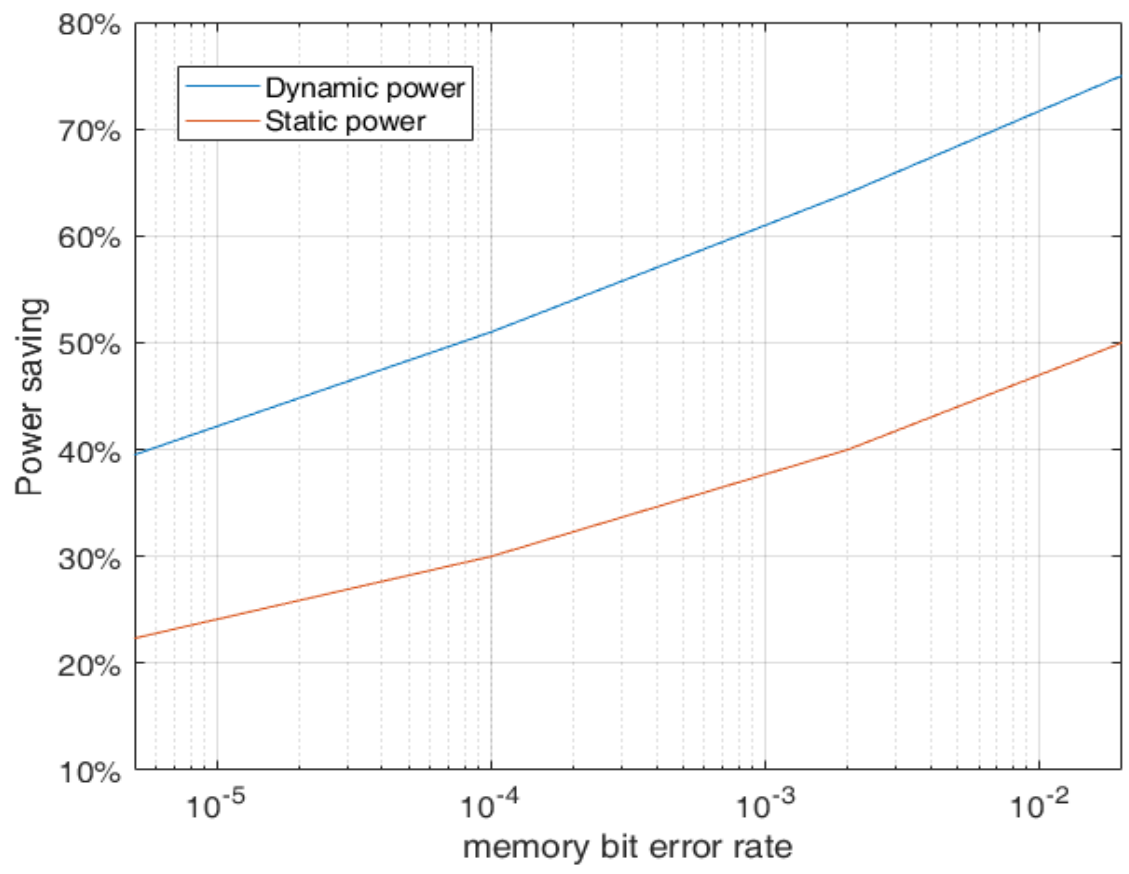

Fig. 4. Power savings for different memory BERs due to supply voltage scaling in a 28 nm approximate SRAM [27].

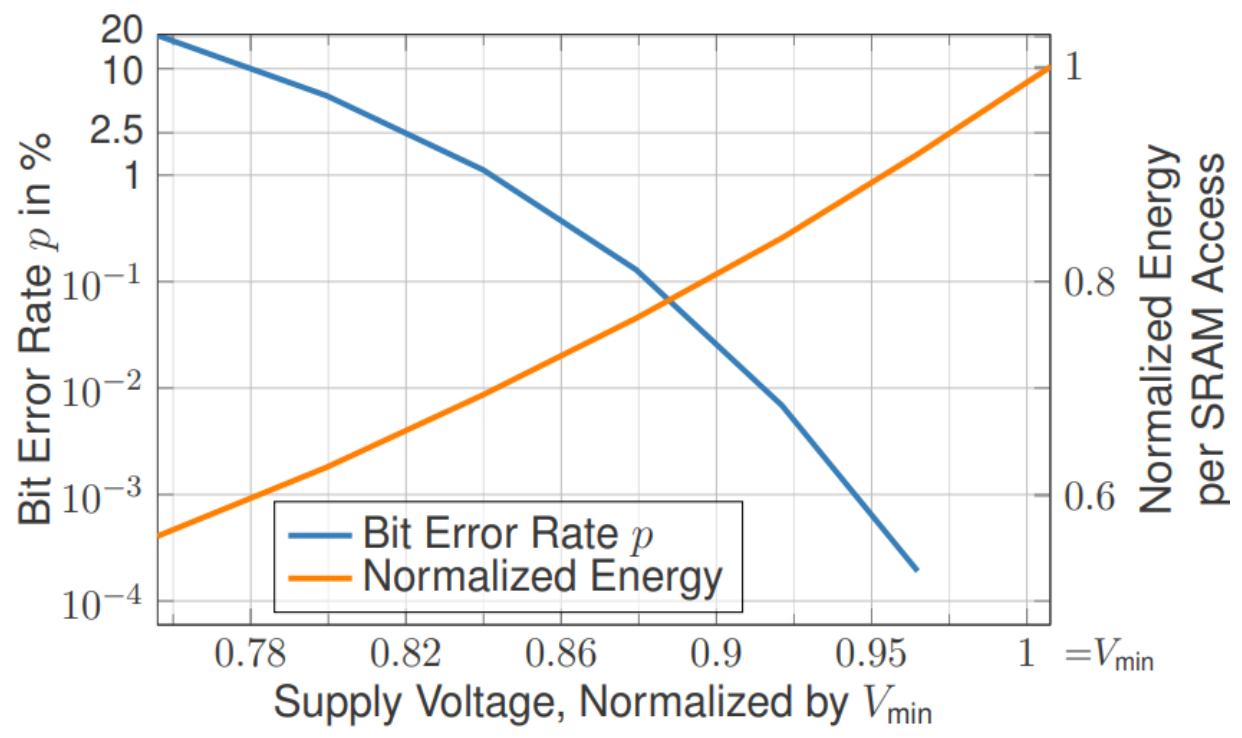

Fig. 5. Relationship between supply voltage, energy saving and BER for SRAM [28].

In this paper, we assume that errors occur on the weights that are stored in persistent storage. For binary quantized weights $\hat{w}$, we assume that the weight matrix only stores the 0/1 bits (for -1 and 1, respectively), and the floating-point scaling factor for each matrix is stored separately (e.g. $\alpha_1$ and $\alpha_2$ in equation (4) and $\|\boldsymbol{W}\|_1/n$ in equation (9)). The errors only flip the 0/1 bits in the weight matrix, and the largest BER is set to be 10% based on the results in [27] ~ [29].

### *B. Effect of errors with different error rates*

In these experiments, we have injected errors on all types of weight matrices in all layers with the BER changing from 0 to 0.1. For each BER, we have repeated the error injection five times and measured the average accuracy for the effect of the errors. The results for the two models are shown in Fig. 6. As shown in Fig. 6(a), the accuracy for BinaryBERT with W1A4 starts to degrade when the BER is larger than 0.05, and it decreases to 66% for a BER of 0.1; the accuracy with W1A8 decreases faster than for a BER larger than 0.02, and the model totally fails at a BER of 0.1 (50% for binary classification). The results in Fig. 6(b) reveal a similar conclusion for BIT. For BER lower than 0.02, performance ranking is W1A8 > W1A4 > W1A1; from 0 BER to 0.1 BER, the accuracy losses for W1A8, W1A4 and W1A1 are 12%, 7% and 3.7%, respectively. Therefore, we can conclude that the model with lower precision activations is more dependable to errors on the binary weights. This is reasonable because larger bit-width for activations implies larger changes of the operation results for the same errors on the weights.

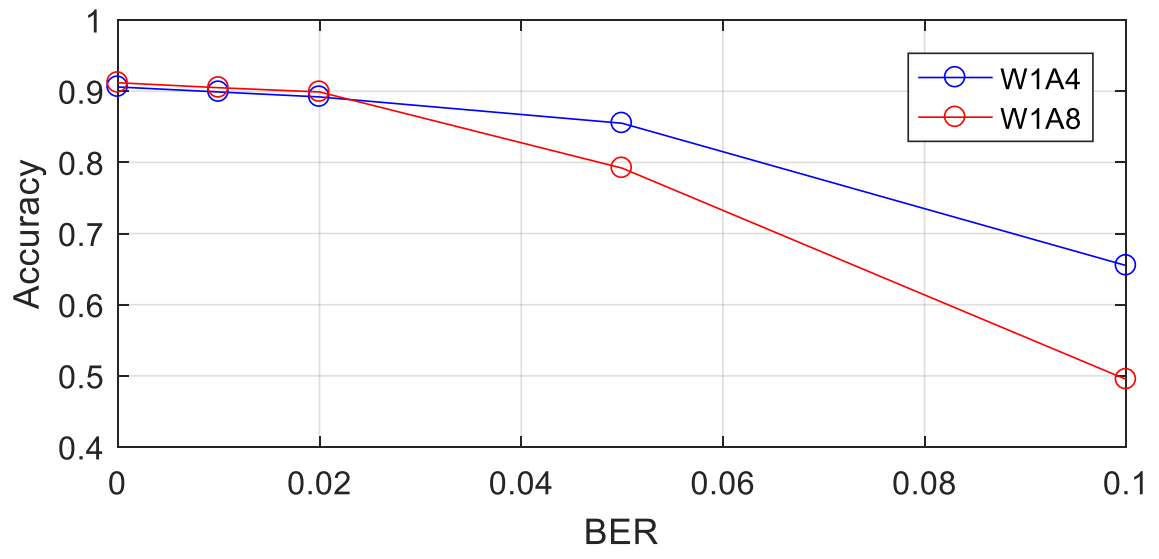

(a) Binary BERT

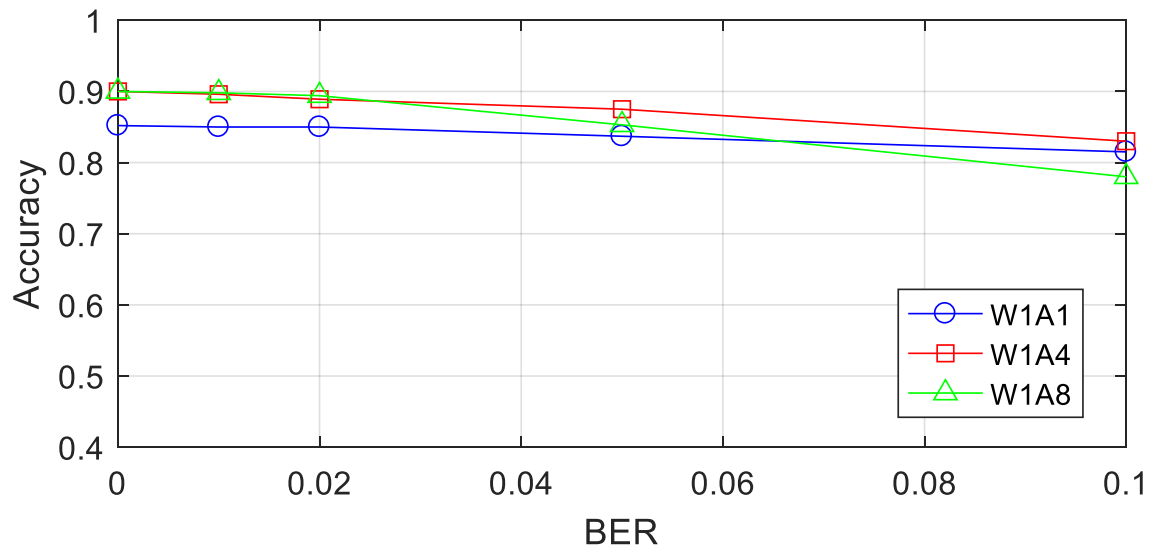

(b) BIT

Fig. 6 Impact of errors on the performance with different BERs

### *C. Effect of errors on different types of weights*

For this assessment, we have injected errors on a specific type of weights in all layers with a fixed BER. To show the dependability of different types of weights, we have selected the high BER of 0.1 for both models based on the results in sub-section B. For each type of weight, the simulation is repeated five times, and the average accuracy for the two models are shown in Fig. 7. A common feature for the two models is that errors on the weights for the first fully connection layer (F1) cause larger accuracy loss than errors on other types of weights. Further analysis shows that the output of F1 ($O_{f1}$) includes most negative values (over 2/3), and all these values would be set to zero by the activation function in the normal case. However, errors on the weights $W_{F1}$ will change some values in $O_{f1}$ from positive to negative (or vice versa), so that many of the positions for the positive values after the activation are changed. Such effect has a larger impact on task performance than only changes for some values of the output of other operations.

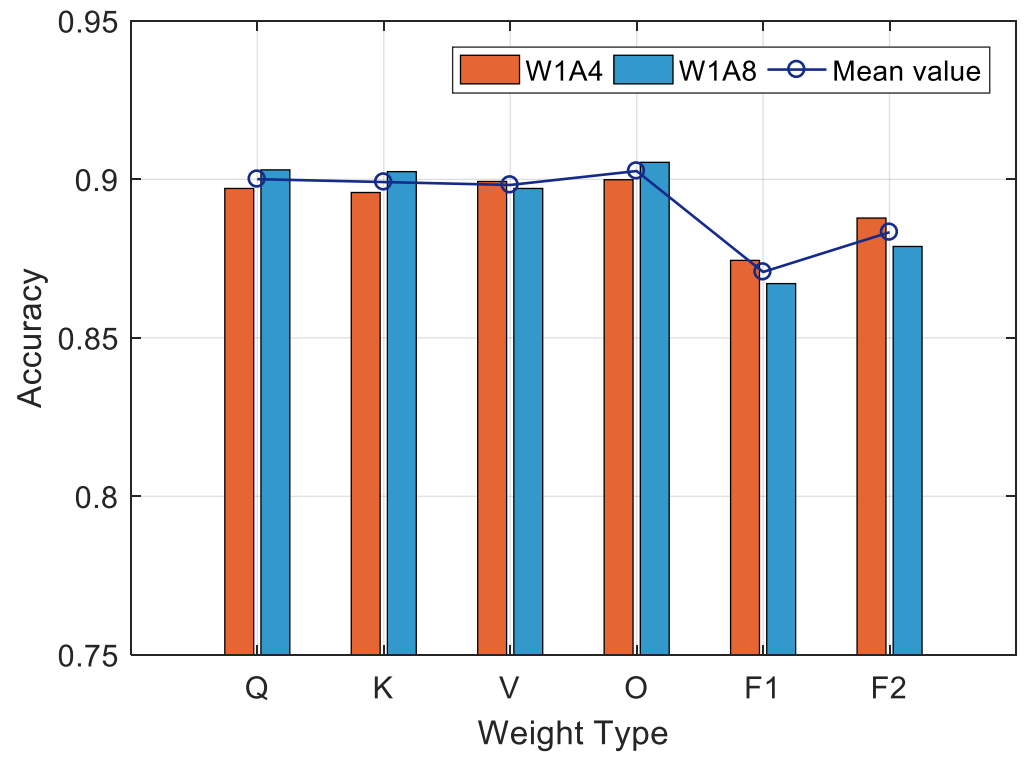

(a) Binary BERT with BER of 0.1

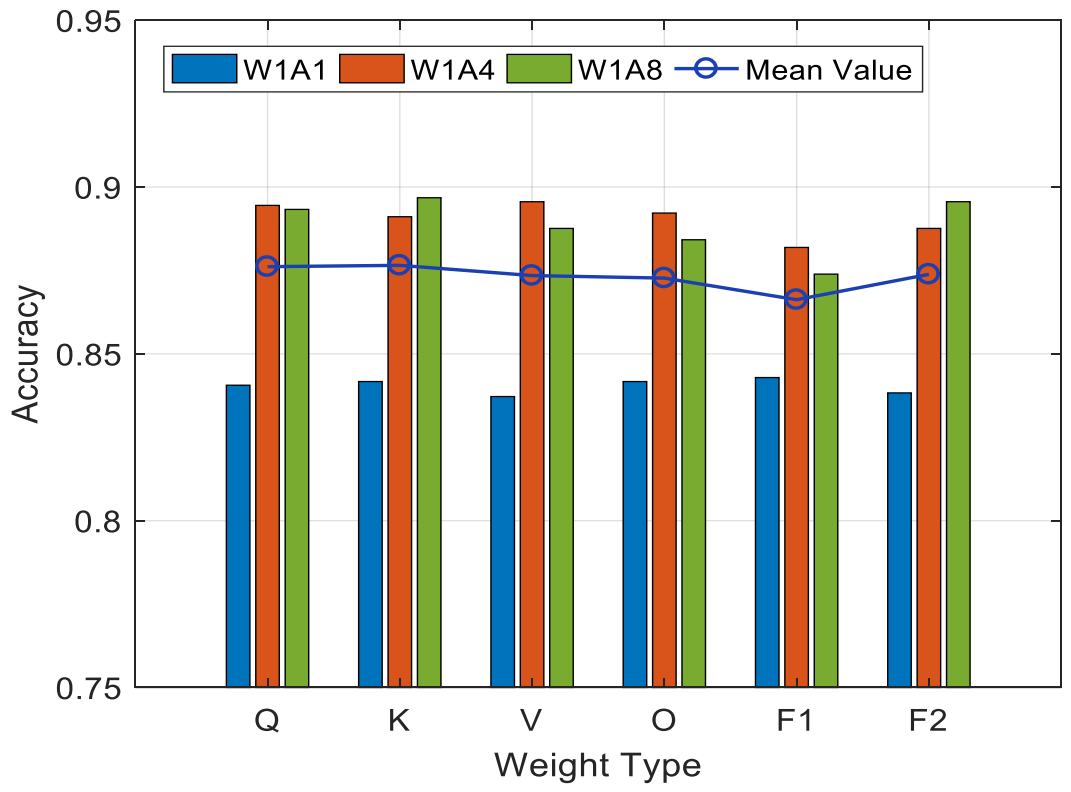

(b) BIT with BER of 0.1

Fig. 7. Effect of errors on different types of weight matrices

## IV. GA Based Critical Weights Identification

In this section, we first introduce briefly the overall approach of the PIES scheme in sub-section A, and then introduce the details of the first step and the evaluation results in sub-sections B and C, respectively.

### A. Overall approach

Since binary BERT shows a significant performance loss only for a BER higher than 5% (as shown in Section III.B), we focus on improving dependability by utilizing a selective protection for high BER. As shown in Fig. 8, the proposed selective protection scheme for binary BERT includes three steps:

*1) Identification of critical weights:* This step includes two parts. The first part sorts the weights in a matrix based on the impact of their flips on the inference loss; the second part searches for the best protection ratio for each weight matrix in a layer based on Genetic Algorithms (GA) [30]. The evaluation results in Section IV. C will show that it is possible to only protect a small portion of critical weights to keep a high performance of binary BERT under a BER as high as 0.1.

*2) Gathering of critical weights:* As shown in Section V.A, the identified critical weights are distributed over the weight matrix, so it is difficult to selectively protect them at a low overhead, because recording their row/column indices would require additional memory. To solve this problem, we apply the permutation invariance property of Transformers to gather the important rows or columns at the beginning of the matrix. Evaluation results in Section V.C will show that such matrix re-organization enables practical selective protection without recording the positions of the critical weights.

*3) Protection of critical weights:* After gathering most of the critical weights in the first few rows and columns, TMR or block ECC (BECC) schemes can be applied to selectively protect them in an uniform fashion. The performance for two case studies is presented in Section VI.

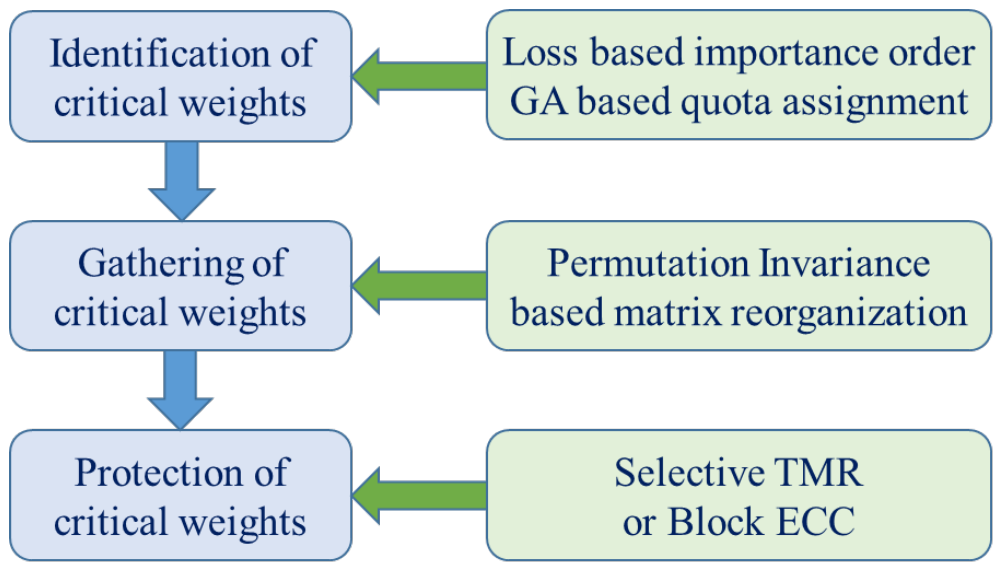

Fig. 8. General procedures of PIES for binary BERT

### B. Identification of Critical Weights

The identification of critical weights for a protection ratio $r$ includes two steps: 1) sort all weights in each weight matrix based on their importance; 2) obtain the best protection ratio for each weight matrix based on a GA method so that the best dependability is achieved for BER of $p$ under the overall protection ratio constraint. The details are included in Algorithm 1, and a more detailed presentation is provided as follows.

***Line 1 ~ Line 6 (Sorting weights in each matrix)***: To evaluate the importance of each binary weight, we first select a small subset of the dataset to calculate the mean loss $L$. Then the loss amplitude caused by the bit flip of the $i$-th binary weight ($w_i$) is derived as

$$|\Delta L_i| \approx 2\alpha \left|\frac{\partial L}{\partial w_i}\right| \tag{12}$$

where $\alpha$ is the scaling factor defined in Eq. (1) and (4). Based on this metric, we sort all the weights in each weight matrix ($W_t^l$, t ∈ {Q, K, V, O, F1, F2}, $l$ = 1, 2, …, 12).

***Line 7 ~ Line 9 (Initialization of population)***: Since the importance of different types of weight in different layers is not the same, we apply GA to determine the number of weights for protection for each weight matrix ($N_t^l$) as

$$\begin{cases} \sum_l \sum_t N_t^l = rN \\ N_t^l \le U_t^l \end{cases}, \tag{13}$$

where $N$ is the total number of weights of the model, and $U_t^l$ is the number of weights in $W_t^l$ ($\sum_{l,t} U_t^l = N$). We define the protection budget as $B = rN$.

In our GA, the population $P$ includes 24 individuals $z_i$ ($i$ = 1, 2, …, 24), and each $z_i$ is a 12*6 matrix, in which the $t$-th item in the $l$-th row is $N_t^l$. The first four individuals are initialized based on four typical strategies {*SizeProp*, *Equal*, *AttenFocus*, *FFNFocus*}. For *SizeProp*, $N_t^l$ is proportional to $U_t^l$; for *Equal*, all $N_t^l$ are the same; for *AttenFocus*, we have $N_{Q/K/V/O}^l = 2N_{F1/F2}^l$; for *FFNFocus*, we have $N_{F1/F2}^l = 2N_{Q/K/V/O}^l$. The other 20 individuals are initialized randomly under the constraint of $N_t^l \le U_t^l$ and $\sum_l \sum_t N_t^l = B$.

***Line 10 ~ Line 18 (Evaluation of each individual)***: The largest generation in GA is set to be 30. For each generation, we first evaluate each individual based on the model accuracy under a BER of $p$. Therefore, we first inject a bit flip error in all weight matrix as in Section III.B (to get $W_f$), and then we recover the $N_t^l$ binary weights in $W_t^l$ for an individual $z$ (to get $W_z$), whose dependability is then evaluated by running the model on the dataset (A(z)). After evaluating all individuals, they are sorted based on their dependability, and the best one (P[1]) is recorded. During the 30 generations, if the best individual keeps the same for 8 generations, the GA stops.

***Line 19 ~ Line 27 (Population evolution)***: The best 2 individuals from P are selected as the parents for the next generation ($P_{next}$). Then the standard Selection→ Crossover → Mutate process is performed to generate the other 22 individuals in $P_{next}$. In our scenario, Crossover produces $z_{child}$ by selecting the value for $N_t^l$ from either $z_a$ or $z_b$; Mutate exchanges the value of some $N_t^l$ for different weights in different layer within $z_{child}$.

After 30 (or less) generations, the best individual is kept as $z^* = \{N_t^l\}$, and the mask matrix $M_{exact} = \{M_t^l\}$ is generated accordingly, i.e., the items for critical weights are 1s and other items are 0s.

**Algorithm 1:** GA-Based Critical Weight Identification

Input: Clean model weights $W$=$\{W_t^l\}$; Dataset $D$; Protection ratio $r$; BER $p$.
Output: Optimal protection-allocation strategy $z^* = \{N_t^l\}$ and the corresponding mask $M_{exact}$ =$\{M_t^l\}$for the exact $N_t^l$ critical weights in $W_t^l$.

1: for $l$ = 1, 1, …, 12 do // *iterate over 12 layers in BERT*
2: for t ∈ {Q,K,V,O,F1,F2} do // *iterate over 6 weight matrices*
3: $G_t^l$ ← MeanAbsGradient($W_t^l$, *D*) // *Gradient for each weight*
4: $SG_t^l$ ← Argsort↓($G_t^l$) // *Sort by Gradient*
5: end for
6: end for
7: S ← {SizeProp, Equal, AttenFocus, FFNFocus} // *typical strategies*
8: P ← [InitByStrategy(s, U, B) | s ∈ S] // *first four individuals*
9: P ← P ∪ RandomIndividualInit(U, B, 24−|P|) // the other 20 by random
10: $W_f$ ← InjectFault(W, p) // *Inject bit flips in all weight matrices*
11: for generation = 1, 2, …, 30 do
12: for each z ∈ P do
13: $W_z$ ← RestoreTopRanked($W_f$, $SG_t^l$, z) // recovery protected
14: A(z) ← Accuracy(Wz,D)
15: end for
16: P ← Sort↓(P, A)
17: UpdateGlobalBest(P[1])
18: if NoImprovement(8) then break
19: $P_{next}$ ← RetainBest(P, 2)
20: while | $P_{next}$ | < 24 do
21: ($z_a$,$z_b$) ← Select($P_{next}$)
22: $z_{child}$ ← Crossover($z_a$, $z_b$)
23: zchild ← Mutate($z_{child}$)
24: Append($P_{next}$, $z_{child}$)
25: end while
26: P ← $P_{next}$
27: end for
30: z* ← GlobalBest
31: $M_{exact}$ ← BuildExactMask(SG, z*)
32: return z*, $M_{exact}$

## C. *Performance for Ideal Protection of Critical Weights*

Based on Algorithm 1, we can determine the most critical weights for a specific protection ratio $r$ and BER. Fig. 9 shows the accuracy for protecting 5% ~ 20% of the critical weights ($r$ = 0.05/0.1/0.15/0.2) under BER of 0.1, in which the accuracy in the error free case and the accuracy for the BER of 0.1 with no protection are provided as dash line and dot line, respectively, for reference. As we can see, protecting only a small portion of the critical weights can improve the dependability of binary BERTs. For example, for the W1A1 version of BIT (left in Fig. 9(a)), protecting 5% of the critical weights almost recovers the accuracy back to the error free case; for the W1A8 version of BIT (right in Fig. 9(a)), the accuracy improvement relative to the unprotected baseline increases from 8% for $r$ = 0.05 to 12% for $r$ = 0.2. The accuracy improvement for BinaryBERT is more obvious (as shown in Fig. 9(b)): for the W1A4 version, the improvement increases from 15% for $r$ = 0.05 to 21% for $r$ = 0.2; for W1A8 version, the improvement for $r$ = 0.05 and 0.2 are 28% and 37%, respectively. These results are for the case that we can ideally protect the identified critical weights that are distributed over the whole weight matrix. Therefore, they are the upper limit for selective protection in practice.

Fig. 10 shows the number of protected critical weights for different types of weight ($N_t^l$ in Algorithm 1) for $r$ = 0.05~0.2 and BER of 0.1. As we can see, over 70% of the critical weights come from $W_{F1}$ and $W_{F2}$. This is reasonable because these two matrices account for 2/3 of all weights (as shown in Table I), and the bit flip errors on $W_{F1}$ and $W_{F2}$ have a higher impact on the accuracy than other types of weight (as shown in Fig. 5). These results show that the proposed GA based Algorithm 1 can accurately identify the important weights for the dependability of binary BERT.

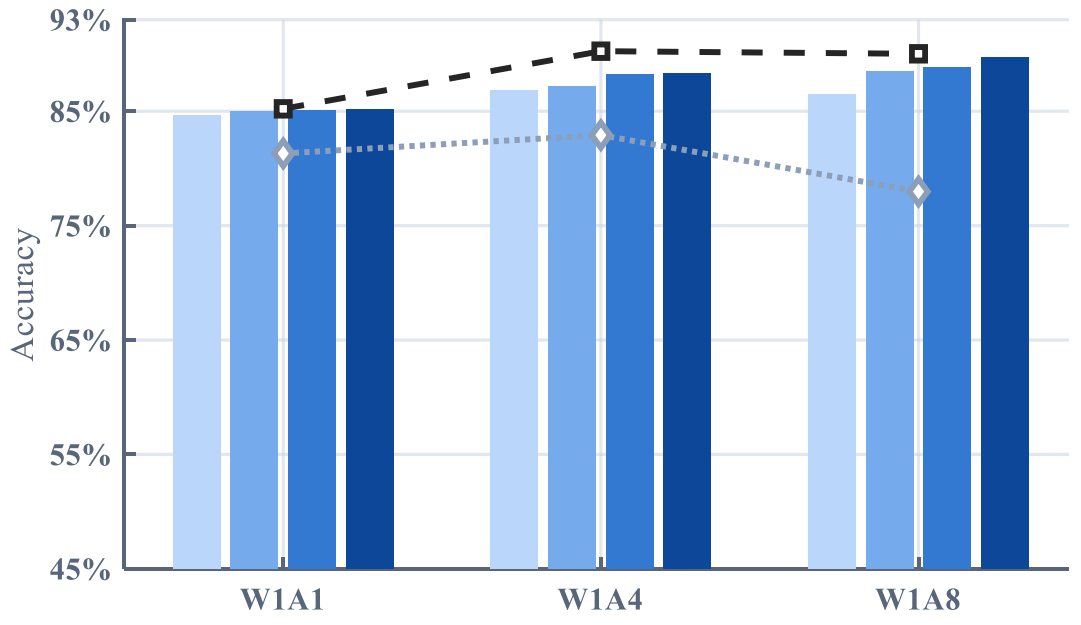

(a) BIT

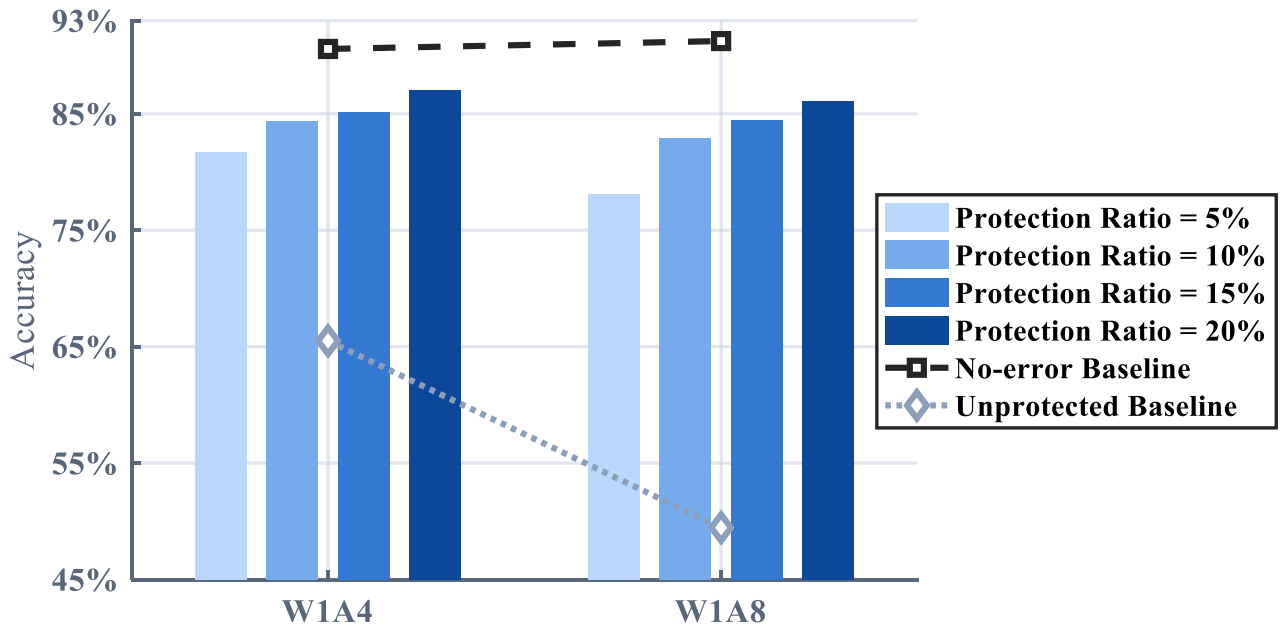

(b) BinaryBERT

Fig. 9. Accuracy for ideal protection of critical weights (BER = 0.1)

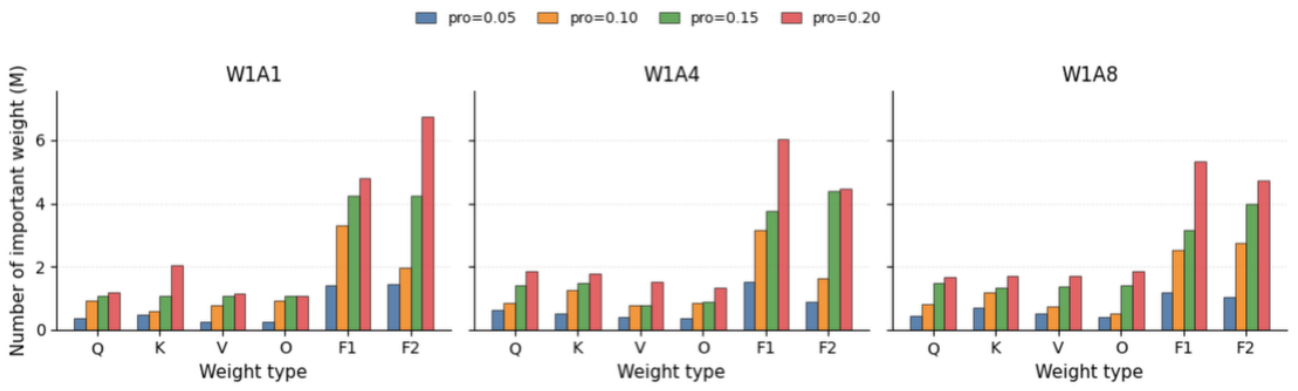

(a) BIT

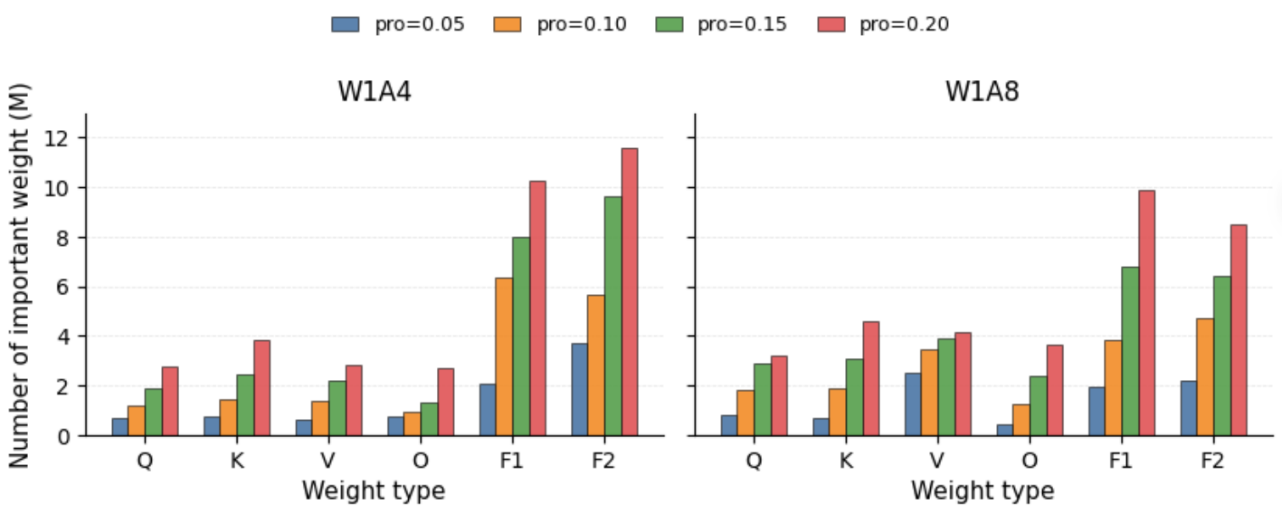

(b) BinaryBERT

Fig. 10. Distribution of critical weights (BER= 0.1)

## V. Permutation Invariance based Gathering of Critical Weights

In this section, we first analyze the distribution of critical weights in weight matrices in sub-section A, and then we propose the GA based protection ratio assignment in sub-section B.

### *A. Distribution Analysis of Critical Weights in Matrices*

Experiment results show that the distributions of critical weights in the Attention module ($W_{Q/K/V/O}$) and FFN ($W_{F1/F2}$) have different properties. Consider the first layer of the BIT (W1A8) as an example, Fig. 11 shows the distribution of 10% critical weights ($r$ = 0.1, BER = 0.1) over the 6 weight matrices, where the light points indicate the critical weights. Based on these figures, several features can be established:

1) The critical weights in all six weight matrices concentrate along the hidden dimension (768).
2) For the Attention module, the critical weights are dominated by several important heads. The width of the light belts in Fig. 11 (a) ~ (d) is 64, which is the head dimension.
3) For a specific head, the inclusion of critical weights may be different for the four weight matrices. For example, for the 2nd and 5th heads, all four weight matrices include a large number of critical bits; for the 7th head, $W_Q$ and $W_K$ include many critical weights; for the 12th head, only $W_K$ include many critical weights.
4) For the FFN weight matrices, the rows or columns with critical weights are sparsely distributed over the whole matrix.

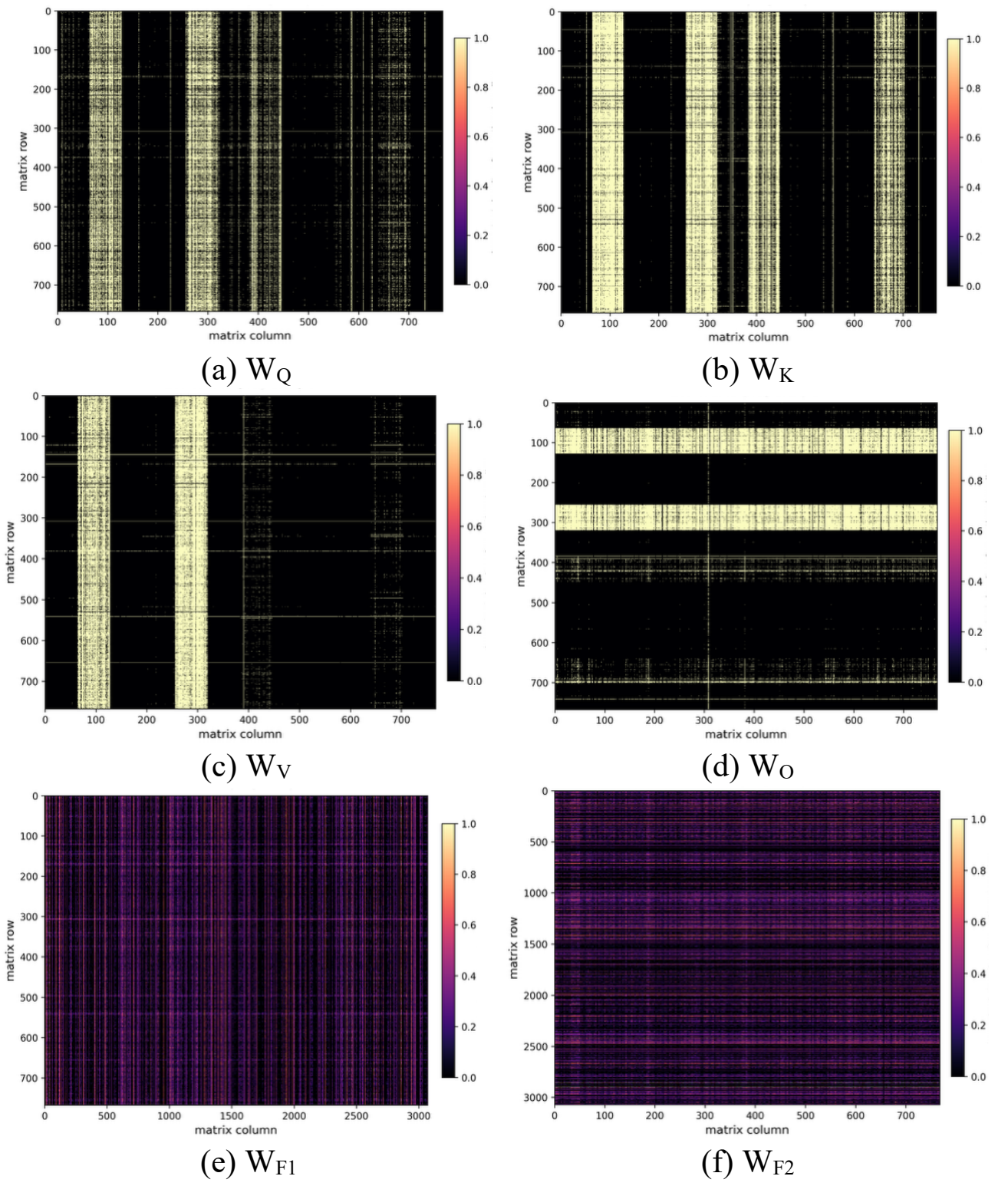

(a) $W_Q$ (b) $W_K$ (c) $W_V$ (d) $W_O$ (e) $W_{F1}$ (f) $W_{F2}$

Fig. 11. Distribution of critical weights in 6 matrices (r = 0.1, BER = 0.1)

### *B. Proposed Approach*

For a selective protection of the critical weights in each matrix, we propose a two-step procedure: 1) gather the critical weights to the first several rows or columns of the weight matrices; 2) obtain the best protection ratio (or number of rows/columns) for each weight matrix based on the GA method. The feasibility for the first step is based on two key findings: 1) as introduced in Section II.D, the permutation invariance holds along the hidden dimension; 2) the first observation in Section V.A reveals that critical weights also concentrate along the hidden dimension. Therefore, it is possible to move the important rows or columns to the front of the weight matrix without changing the inference result of the binary BERTs.

Algorithm 2 is designed for the two steps, and more explanation is given as follows.

***Line 1 ~ Line 10 (Weight matrix permutation based on importance of rows/columns)***: Based on the positions of the critical weights in each weight matrix ($M_{exact}=\{M_t^l\}$), the importance for each row or column in $W_t^l$ is measured according to the number of included critical weights. For the Attention part, based on the 2nd observation in Section V.A, we first sort all heads in a layer based on the total number of critical weights in the four weight matrices ($c_Q$, $c_K$, $c_V$, $c_O$). In each head, to force synchronous permutation for each PI pair, we sort each row in the Q/K PI pair based on the sum of the importance of the corresponding row ($c_{QK} = c_Q + c_K$), and sort the columns/rows in the O/V PI pair based $c_{OV} = c_V + c_O$. For the FFN part, we sort the columns/rows in the F1/F2 PI pair based $c_{FFN} = c_{F1} + c_{F2}$. Then we perform the matrix permutation for each PI pair based on the sorted rows/columns, and obtain the new weight matrix $W_{perm}$ and the corresponding mask for critical weights $M_{perm}$.

***Line 11 ~ Line17 (Initialization of population)***: The protection budget (total number of rows or columns) is determined as $B = ||M_{perm}||_0/b$, where $||M_{perm}||_0$ is the number of 1s in $M_{perm}$, and $b$ = 768 is the hidden dimension. We then apply GA to determine the number of rows/columns for protection for each weight matrix ($R_t^l$) under the budget as

$$\begin{cases} \sum_l \sum_t R_t^l = B \\ R_t^l \le U_t^l \end{cases}, \tag{13}$$

where $U_t^l$ is the number of rows/columns in $W_t^l$.

Same as Algorithm 1, the population $P$ includes 24 individuals $z_i$ ($i$ = 1, 2, …, 24), and each $z_i$ is a 12*6 matrix, where the $t$-th item in the $l$-th row is $R_t^l$. The first three individuals are initialized based on three typical strategies {*OverallRowSort*, *CriticalProp*, *SizeProp*} under the constraint of $U = \{U_t^l\}$ and $B$. For *OverallRowSort*, all rows or columns in all weight matrices are sorted based on the number of included critical weights, and $R_t^l$ is set as the number of rows/columns within the first B ones. For *CriticalProp* and *SizeProp*, $R_t^l$ is proportional to $c_t^l$ and $U_t^l$, respectively. The other 21 individuals are initialized as the mutation of the first individual (P[1] by *OverallRowSort*).

***Line 18 ~ Line 27 (Evaluation of each individual)***: The largest generation is still 30. For each generation, we first evaluate each individual based on the model accuracy under BER of $p$ following the same method in Algorithm 1 (A(z)), and then we get the coverage of critical weights by z (C(z)). All individuals are sorted based on A(z) and C(z), and the best one (P[1]) is recorded. Same as in Algorithm 1, the iteration stops if the best individual keeps the same for 8 generations.

***Line 28 ~ Line 38 (Population evolution)***: The evolution process is the same as in Algorithm 1. After 30 (or less) generations, the best individual is kept as $z^* = \{R_t^l\}$, and the weight matrices after permutation are protected by TMR or ECC as in Section VI.

**Algorithm 2**: Permutation invariance based weight matrix reorganization and GA based protection ratio assignment

Input: Clean weights W, exact mask $M_{exact}$ for critical weights obtained in Section IV.B, dataset D, BER p = 0.1 and row size $b$ = 768.
Output: Permuted weights $W_{perm}$, optimal protection stratigies z* = $\{R_t^l\}$.

```
for l = 1, 2, …, 12 do
    (c_Q^l, c_K^l, c_V^l, c_O^l, c_F1^l, c_F2^l) ← CriticalWeightsCount(M_exact, l)
    πh ← Argsort↓(HeadSum(c_Q^l + c_K^l + c_V^l + c_O^l))
    for h ∈ πh do
        πqk(h) ← Argsort↓(c_Q^l(h) + c_K^l(h))
        πvo(h) ← Argsort↓(c_V^l(h) + c_O^l(h))
    end for
    πffn ← Argsort↓(c_F1^l + c_F2^l)
    (W_perm, M_perm) ← PermuteSync(W, M_exact, πh, πqk, πvo, πffn)
end for
B ← round(|| Mperm ||_0/b)
S ← {OverallRowSort, CriticalProp, SizeProp} // typical stratigies
P ← [InitByStrategy(s, U, B) | s ∈ S] // first 3 individuals
while |P| < 24 do // initiate other 21 individuals
    z ←Mutate(P[1], U)
    Append(P, z)
end while
W_f ← InjectFault(W_perm, p)
for generation = 1, 2, …, 30 do
    for each z ∈ P do
        Wz ← RestorePrefix(W_f, z)
        A(z) ← Accuracy(Wz, D)
        C(z) ← ||M_perm(z)||_0
    end for
    P ← Sort↓(P, (A,C)) // Compare C if A is the same
    UpdateGlobalBest(P[1])
    if NoImprovement(8) then break
    P_next ← RetainBest(P, 2)
    while |P_next| < 24 do
        (z_a, z_b) ← Select(P)
        z_child ← Crossover(z_a, z_b)
        z_child← Mutate(z_child)
        Append(P_next, z_child)
    end while
    P ← P_next
end for
z* ← GlobalBest
return W_perm, z*
```

### C. *Impact of Permutation on Protection Performance*

This section reports the performance of binary BERTs under BER of 0.1 when $M_t^l$ rows/columns of each weight matrix are ideally protected. This would be the upper limit for practical implementation of PIES based on TMR or ECC.

Consider still the first layer of the BIT (W1A8) as an example, Fig. 12 shows the distribution of 10% critical weights ($r$ = 0.1, BER = 0.1) over the 6 weight matrices after gathering the critical weights based on Algorithm 2. By comparing it with Fig. 10, we can find that most of the critical weights are moved to the first several rows or columns, but there are still some important rows or columns (*Outliers*) that are not covered by the first rows/columns for the Attention part. This introduces some performance loss relative to the ideal case of Section IV.C.

We define the performance retention rate of PIES as

$$R_{\text{acc}}(\text{r}) = \frac{Acc_{\text{prefix}}(r)}{Acc_{\text{exact}}(r)} \times 100\%, \qquad (14)$$

where $Acc_{\text{prefix}}(r)$ denotes the accuracy of the model when $R_t^l$ rows/columns at the front (*prefix*) of each weight matrix are ideally protected ($768^*\sum_{l,t} R_t^l = rN$), and $Acc_{\text{exact}}$(r) denotes the accuracy when the exact $rN$ critical weights are protected ideally (as shown in Fig. 9). Fig. 13 shows the result of $R_{acc}$ for different protection ratio and different versions of the two binary BERTs. The value of $R_{acc}$ is higher than 99% for most of the cases; the lowest value is 98.32% for the case of BIT-W1A8 with $r$ = 0.1. These results show that the proposed PIES gathers most of the critical weights, and the few *outlier* rows/columns introduce a negligible performance loss. Therefore, we can focus on the protection of the first $M_t^l$ rows or columns to improve the dependability of the model.

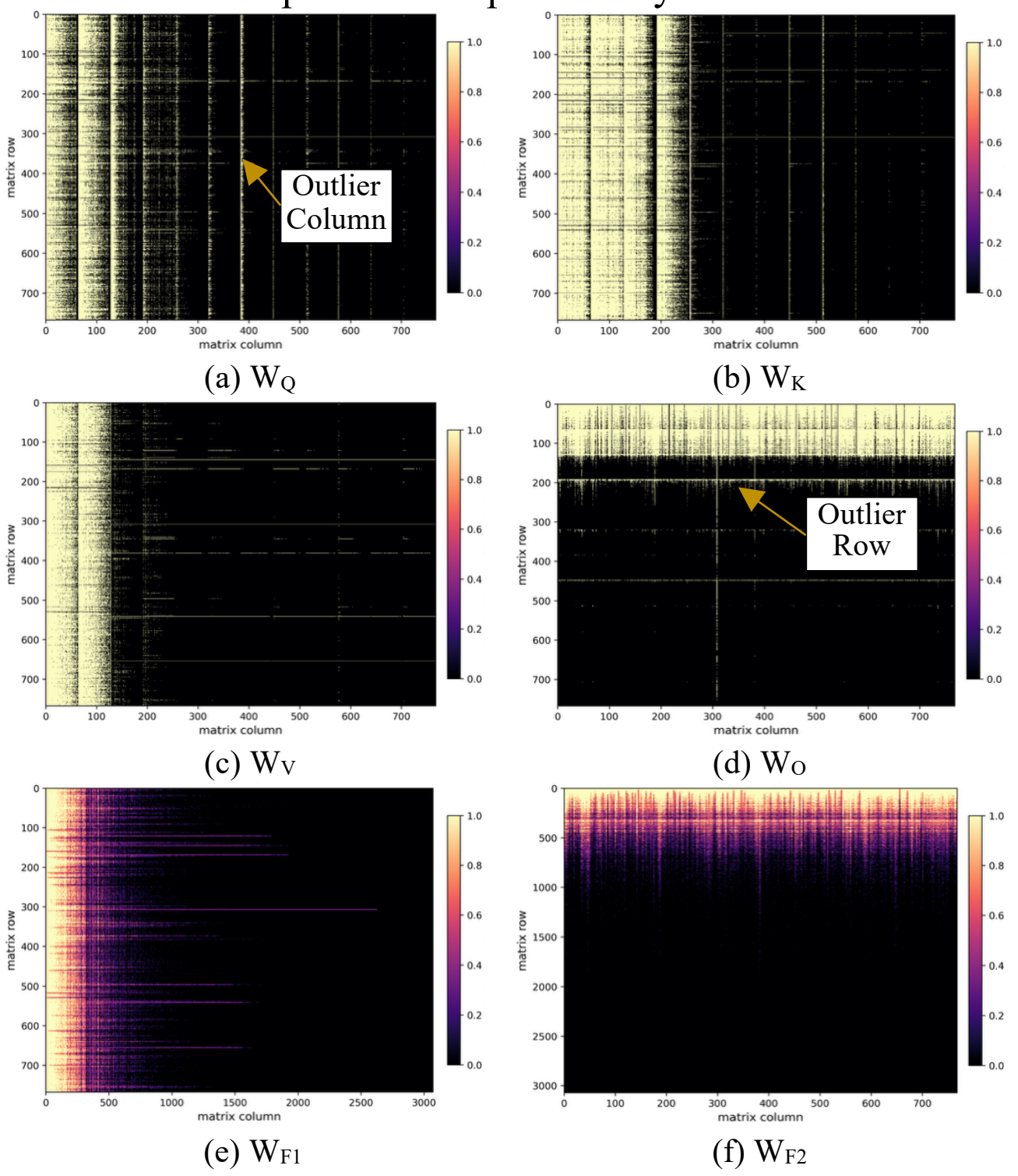

Fig. 12. Distribution of critical weights after permutation (r = 0.1, BER = 0.1)

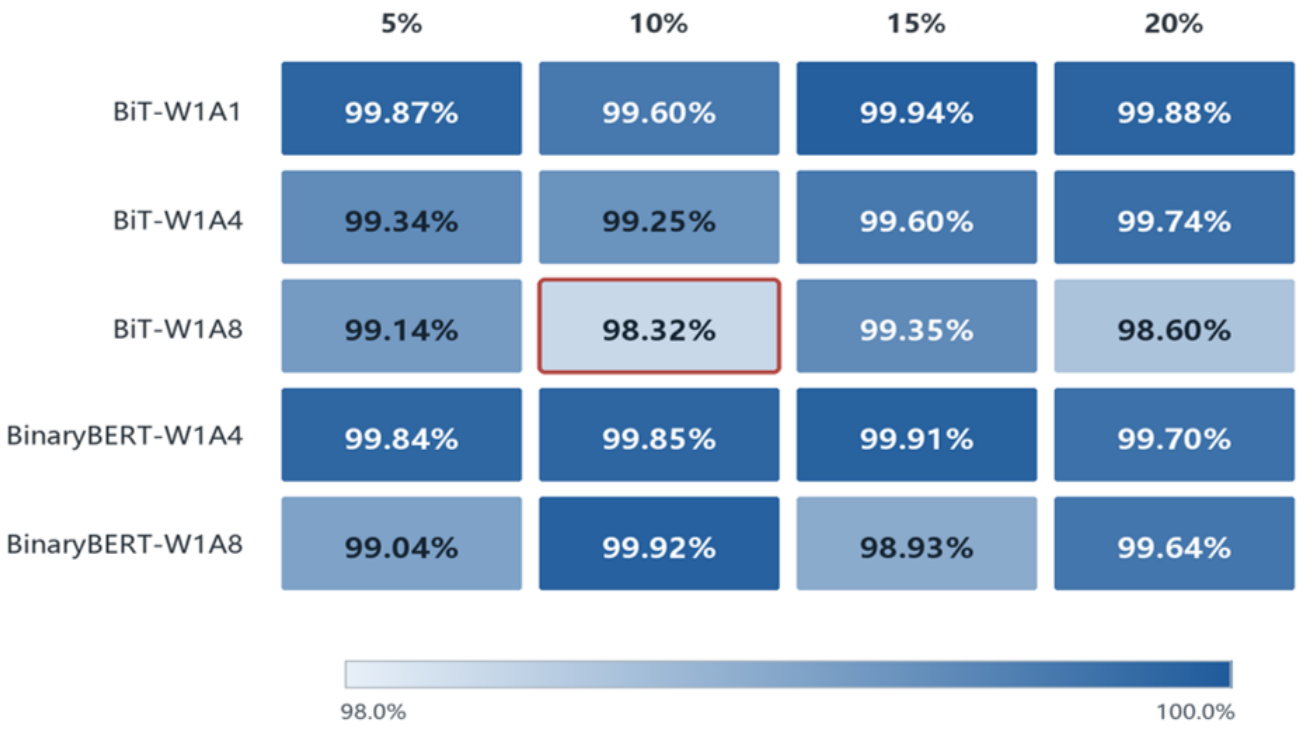

Fig. 13. Accuracy Retention rate of PIES for different protection ratios

## VI. Practical Implementation of PIES

In this section, we first present the results for two PIES implementations (selective TMR and Block ECC) in sub-section A and B, respectively, and then compare them in terms of overhead for different schemes in sub-section C.

### A. *PIES with Selective TMR (PIES-TMR)*

The complete PIES-TMR scheme is implemented for the protection of critical weights in the two binary BERTs for the target BER of 0.1. We first locate the critical weights based on Algorithm 1 for protection ratios ($r$) of 5%, 10%, 15% and 20%. For each value of $r$, we reorganize the weight matrices based on Algorithm 2 and get the value of $M_t^l$ for each weight matrix. Then we apply selective TMR to protect the $M_t^l$ rows or columns: we replicate the $M_t^l$ rows or columns three times and use a majority voter to determine the value of the binary weights during inference. Therefore, the memory of PIES-TMR can be denoted as 1+2$r$. We apply two conventional protection schemes as references, including a (7, 4) Hamming

Code for the entire weight matrix (Full HC) and TMR for the entire weight matrix (Full TMR), whose memory overheads are 1.75 and 3, respectively.

Then we inject bit flip errors on all weight matrices with BER of 0.1 as in Section III.B and check the dependability of the PIES-TMR scheme. The results for a different protection ratio $r$ are shown in Fig. 14 for the two binary BERTs, in which the results for Full HC and Full TMR are denoted by blue circles and red circles, respectively. The main conclusions are summarized as follows.

- For BIT, the advantage of PIES-TMR is not obvious: Full-HC is better in most cases, and PIES-TMR with $r = 0.2$ outperforms the Full-HC scheme by about 0.7% and 0.6% for W1A4 and W1A8, respectively. In particular, the performance for W1A4 with $r = 0.2$ approaches the Full TMR, but the overhead is significantly lower (1.6 vs 3).
- For BinaryBERT, the advantage of PIES-TMR over Full HC is rather obvious: for W1A4, PIES-TMR with r = 0.1 achieves the same performance as Full HC but with a significantly lower memory overhead (1.3 vs 1.75), and outperforms Full HC by 1.7% and 3.8% for $r = 0.15$ and $r = 0.2$, respectively; for W1A8, PIES-TMR outperforms Full HC for all the four protection ratios, and the improvements are 1.5%, 13.2%, 15.3% and 16.8%, respectively. Moreover, the gap between PIES-TMR and Full TMR is about 5%.

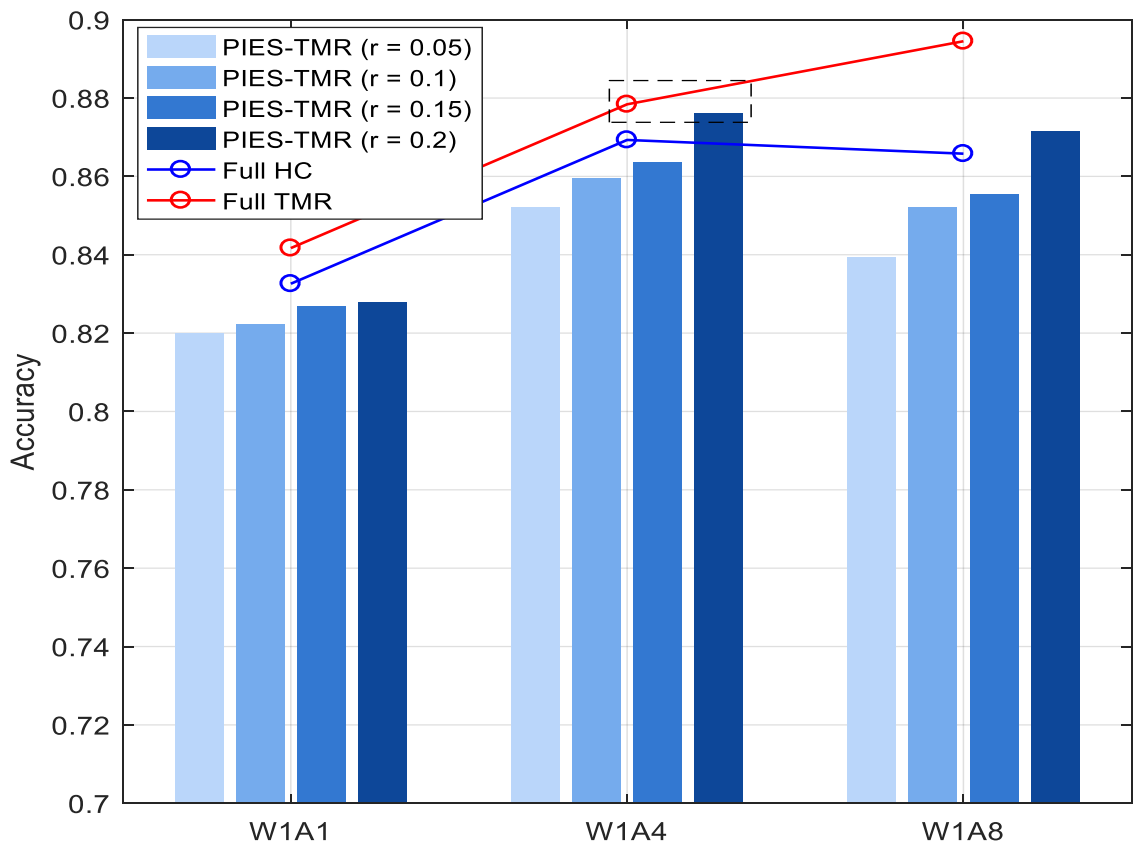

(a) BIT

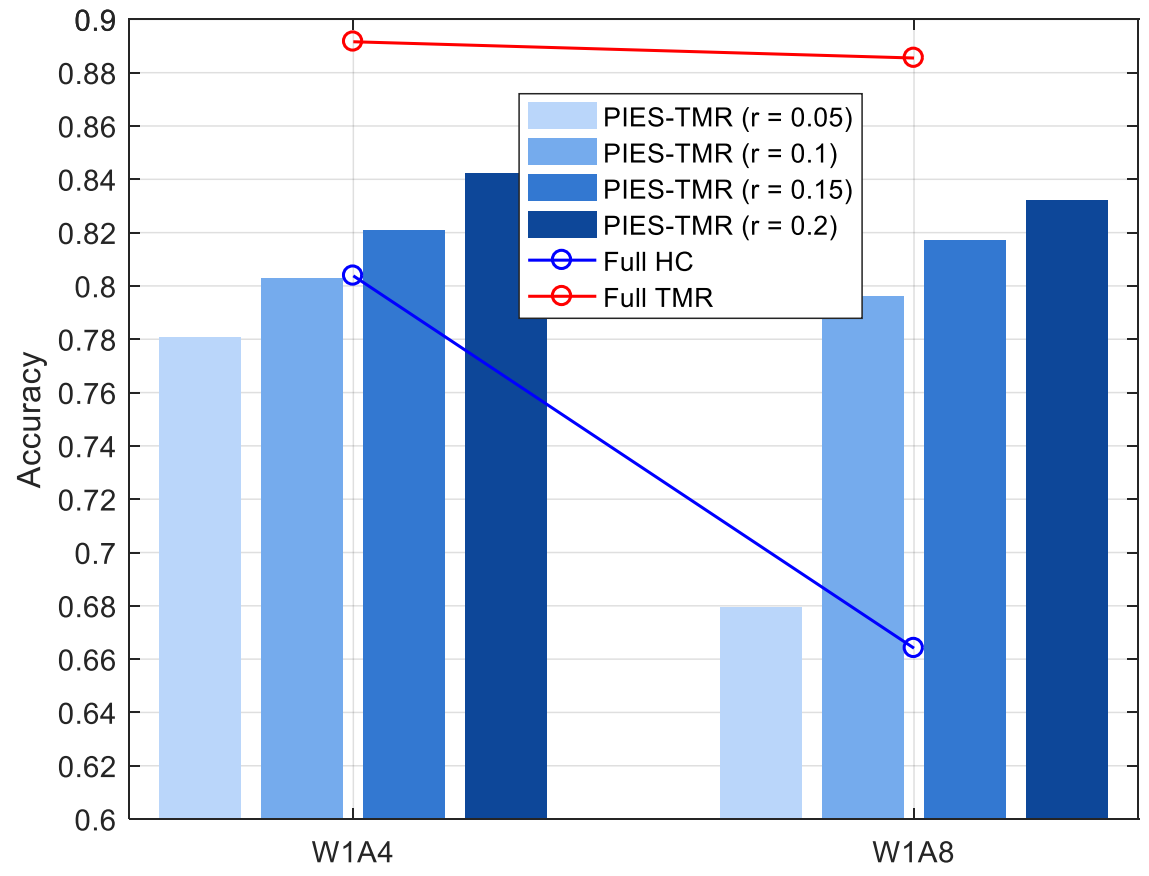

(a) BinaryBERT

Fig. 14. Performance of PIES-TMR for different protection ratio $r$

## B. PIEC with Block ECC (PIEC-BECC)

### 1) Design of the Block ECC scheme

After determining the value of $M_t^l$ for each weight matrix based on Algorithm 2, different BECC schemes can be applied to protect them. We give a reference design in this section based on the sparse graph model and Belief Propagation (BP) based iterative decoding for Low-Density Parity-Check Codes (LDPC) [32]. The basic procedure is introduced as follows (steps a~d for encoding, step e for decoding):

a) For a row or column with 768 items (bits), we divide it into 3 segments with 256 bits and reshape each segment into a 16*16 block.

b) For each block, each row and each column are divided into $m$ vectors with $n$ items ($m = 16/n$);

c) A checksum bit is generated for each vector, so that we get $16m$ row checksum bits and $16m$ column checksum bits;

d) 256 cross checksum bits with degree of 4 are generated based on the sparse graph model of LDPC, in which the $h$-th checksum bit ($h = 0, 1, \ldots, 255$) is the XOR of four data in the block that are determined based on Eq. (15);

$$\Phi(h) = \{17h, 29h + 43, 47h + 91, 71h + 137\} mod 256 \quad (15)$$

e) For decoding, the sparse graph is first built based on the block data and the checksum bits (with errors), then the flipped bits are recovered by 20 BP iterations.

### 2) Performance Evaluation of PIES-BECC

The complete PIES-BECC scheme is implemented for BER of 10% to show the protection performance. In this case study, we set $m = n = 4$ so the total number of checksum bits for each segment (256 bits) is 384. In this case, the overall memory overhead of the BECC can be expressed as a function of $r$ as

$$S(r) = (1 - r) + \frac{640}{256} r = 1 + 1.5r. \quad (15)$$

The results for different protection ratio $r$ are shown in Fig. 15 for the two binary BERTs, in which the points for $r = 0$ correspond to the case with no protection; the results for the Full HC scheme and the PIES-TMR are also provided for comparison. Hence, the following deductions can be drawn.

- For BIT, PIES-BECC with $r = 0.1$ has a performance similar to the Full HC scheme, and the memory overhead is reduced from 75% to 15%;
- For BinaryBERT, even PIES-BECC with $r = 5\%$ is significantly better than the Full HC scheme, and the memory overhead is reduced from 75% to 7.5%.
- Compared with PIES-BECC, PIES-TMR has similar performance with additional 10% ~ 20% memory overhead, but its computation complexity is significantly lower.

These results show that the proposed PIES scheme can effectively improve the dependability of binary BERTs at a high BER with a very low memory overhead. The BECC scheme for other protection ratio and BER can be designed with a similar approach but at a different overhead.

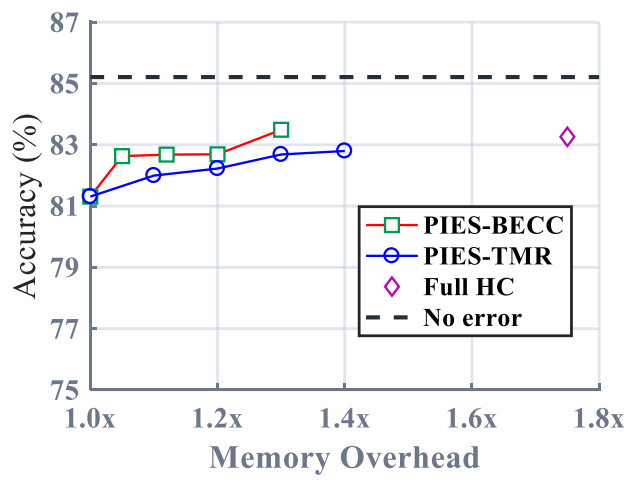

(a) BIT-W1A1

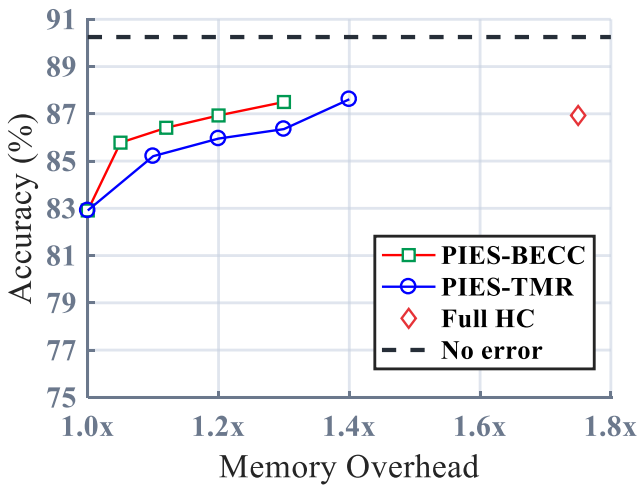

(b) BIT-W1A4

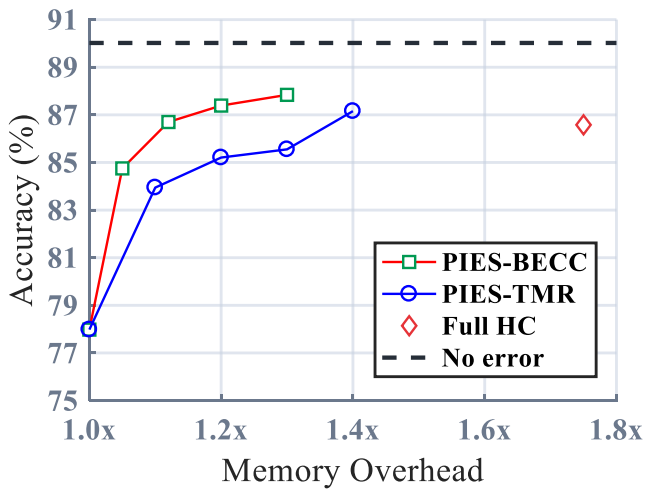

(c) BIT-W1A8

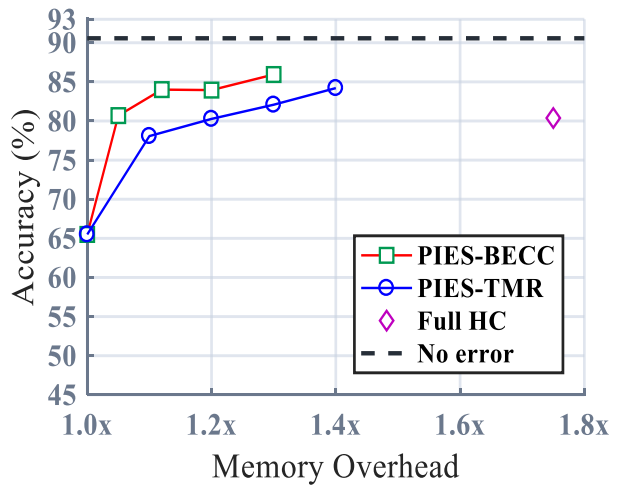

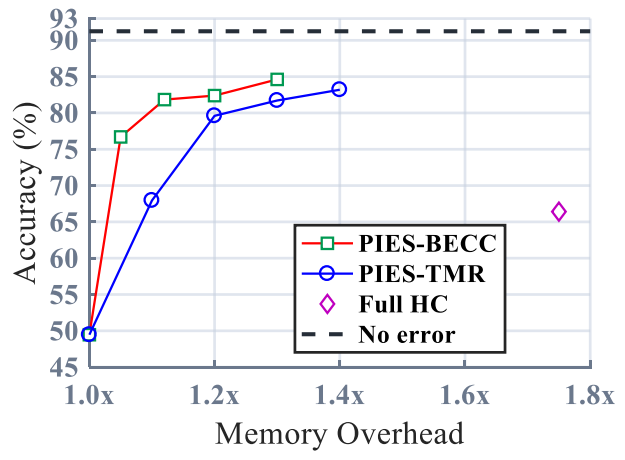

(d) BinaryBERT-W1A4 (e) BinaryBERT -W1A8

Fig. 15. Performance of PIES-BECC for different protection ratio *r*

## VII. Conclusion

Binary Transformers use binary weights to reduce memory storage, inference complexity and power consumption. Using approximate memory can further reduce energy dissipation but it introduces errors due to bit flips. For the deployment of binary Transformers with approximate memory, this paper initially performed a complete dependability evaluation for two typical binary BERTs against bit errors and then an efficient selective protection scheme has been proposed. The evaluation results show that binary BERTs are very dependable to bit flip errors on binary weights, such errors can only degrade for BERs higher than 5%. Then we have proposed a Permutation Invariance Enabled Selective (PIES) protection scheme to improve the dependability at high BER with low overhead. PIES includes three steps: a) a GA based approach is applied to identify the most critical weights; b) since the identified critical weights gather in rows or columns along the hidden dimension, which match the dimension of permutation invariance (PI) in Transformers, we have proposed to gather the rows or columns with large number of critical weights at the front of the matrices based on PI and apply a GA approach to assign the best protection ratio for each weight matrix; c) we have proposed to protect the first several rows/columns with TMR and block ECC and show that the PIES scheme can effectively improve the dependability of binary BERTs with a significantly lower memory overhead than a scheme with complete protection with ECC. The basic principles of PIES scheme can also be applied for pruning or compressing non-binary Transformers or LLMs.